%% file: main.tex
\documentclass[10pt,twocolumn,letterpaper]{article}

\usepackage[pagenumbers]{cvpr} 

\input{preamble}
\definecolor{cvprblue}{rgb}{0.21,0.49,0.74}
\usepackage[pagebackref,breaklinks,colorlinks,allcolors=cvprblue]{hyperref}
\usepackage{multirow}
\usepackage[accsupp]{axessibility}  
\def\paperID{7} 
\def\confName{CVPR}
\def\confYear{2026}

\title{Multiple Domain Generalization Using Category Information \\Independent of Domain Differences}

\author{
Reiji Saito, and
Kazuhiro Hotta\\
Meijo University, 
1-501 Shiogamaguchi, Tempaku-ku, 
Nagoya 468-8502, Japan\\
{\tt\small 200442065@ccalumni.meijo-u.ac.jp, kazuhotta@meijo-u.ac.jp}
}

\begin{document}
\maketitle
\input{sec/0_abstract}    
\input{sec/1_intro}
\input{sec/2_related}
\input{sec/3_method}
\input{sec/4_experiments}
\input{sec/5_conclusion}
{
    \small
    \bibliographystyle{ieeenat_fullname}
    \bibliography{main}
}


\end{document}

%% file: sec/0_abstract.tex
\begin{abstract}
Domain generalization is a technique aimed at enabling models to maintain high accuracy when applied to new environments or datasets (unseen domains) that differ from the datasets used in training. Generally, the accuracy of models trained on a specific dataset (source domain) often decreases significantly when evaluated on different datasets (target domain). This issue arises due to differences in domains caused by varying environmental conditions such as imaging equipment and staining methods. Therefore, we undertook two initiatives to perform segmentation that does not depend on domain differences. We propose a method that separates category information independent of domain differences from the information specific to the source domain. By using information independent of domain differences, our method enables learning the segmentation targets (e.g., blood vessels and cell nuclei). Although we extract independent information of domain differences, this cannot completely bridge the domain gap between training and test data. Therefore, we absorb the domain gap using the quantum vectors in Stochastically Quantized Variational AutoEncoder (SQ-VAE).
In experiments, we evaluated our method on datasets for vascular segmentation and cell nucleus segmentation. Our methods improved the accuracy compared to conventional methods.
\end{abstract}

%% file: sec/1_intro.tex
\section{Introduction}
\label{sec:intro}

Semantic segmentation is a technique for classifying images at the pixel level and is applied in various fields such as medical imaging~\cite{HRNet,V-Net}, autonomous driving~\cite{ESVSNet}, and cellular imaging~\cite{L-UNet,F-UNet}.
Conventional methods~\cite{onepeace,mask2former} are typically trained on specific datasets (source domains) and evaluated on the same datasets. However, these methods often perform poorly when evaluated on different datasets (target domains) due to domain shift.
In medical segmentation, domain shift is particularly pronounced because images are captured in various hospitals and clinical settings.
Domain shift occurs due to differences in imaging conditions, such as imaging devices, lighting, and staining methods. Ideally, accuracy should be maintained regardless of the dataset used for evaluation. Addressing this domain shift and effectively extracting category information that is independent of these differences is a long-standing challenge in deep learning.

One common approach to solving the domain shift problem is domain adaptation (DA). DA leverages labeled data from the source domain to adjust its distribution to match the target domain, maximizing performance on the target domain. However, this approach requires capturing and learning from target domain images, which can be time-consuming. Additionally, DA is only applicable to the specific target domain images being trained, lacking generalizability. Furthermore, in segmentation tasks, manual annotation is required, which can be a significant burden for researchers.

Domain generalization (DG) has been proposed to address the limitations of DA. DG leverages only the source domain to extract features that are not specific to it (e.g., cell nuclei and blood vessels), thereby mitigating domain shift when encountering unseen target domains. Here, we focus on developing a model that effectively generalizes across diverse medical imaging conditions, enhancing robustness and adaptability to varying environments.
Research on DG~\cite{robustnet,IBN} has developed methods that eliminate domain-specific style information from images and use content information for learning. Specifically, these methods involve whitening style-specific features based on the correlation of feature values, thereby retaining content information and improving generalization performance. However, WildNet~\cite{wild} noted that style information also contains essential features for semantic category prediction, and addressing this issue has been reported to improve accuracy.

To address DG without removing style information, we have employed the following two approaches. First, we proposed a method to split feature maps into two parts: domain-invariant category information and source domain-specific information. Specifically, we divide the feature maps along the channel dimension and use DeepCCA~\cite{DeepCCA} to decorrelate these parts. DeepCCA maximizes the correlation between two variables, but we train it to make the correlation zero to extract source domain-specific information and domain-invariant category information.
We train one of the split feature maps to represent domain-invariant category information. Specifically, we train the feature vectors of the same category, based on the ground truth labels, to approach a learnable representative vector. Since the two feature maps are decorrelated, the other feature map becomes the source domain-specific information. We use the obtained domain-invariant category features for segmentation.
Second, although we extracted domain-invariant category information, this alone cannot completely prevent the domain gap between the source and target domains. Therefore, we propose a method to mitigate the domain gap using quantum vectors from SQ-VAE~\cite{SQVAE}. SQ-VAE is a method for reconstructing high-resolution input images, capable of representing images using only quantum vectors.

\begin{figure*}[t]
    \centering
    \includegraphics[width=0.9\linewidth]{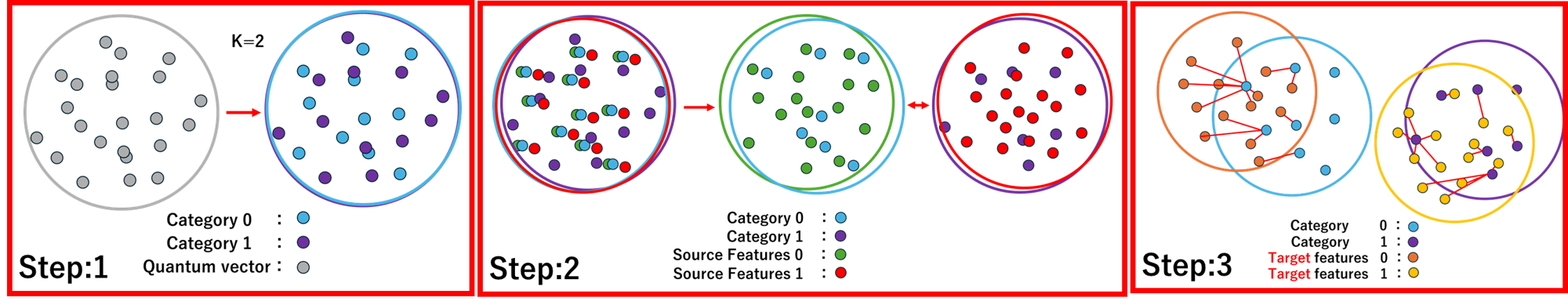}
    \caption{Overview of domain generalization using quantum vectors. This figure explains learning method for quantum vectors used to absorb domain gaps and method to handle unseen target domains during inference.}
    \label{fig:domaincodebook}
\end{figure*}

Figure \ref{fig:domaincodebook} shows the overview of DG using SQ-VAE. In Step 1, we divide N quantum vectors into K groups, where N represents the number of quantum vectors and K represents the number of categories. When an input feature is assigned to the quantum vector defined as category 0, the feature is predicted as category 0. In Step 2, we use ground truth labels to group the features 
into K categories and train the model to bring these groups of the same category closer together. Step 3 is inference. 
Since we do not have access to the target domain, a domain gap arises. However, by aligning the groups of each category, we can minimize the domain gap. Even if there is a gap in the unseen target domain, the features of the target domain are likely to be assigned to the same or similar quantum vectors as those in the source domain and thus categorized similarly.
By using these methods, we can prevent accuracy degradation due to unseen target domains.

Experiments were conducted to segment blood vessels from retinal image datasets (Drive~\cite{Drive}, Stare~\cite{Stare1}, Chase~\cite{Chase}). Each dataset has a different domain due to varying imaging devices. Two retinal image datasets were used for training, while the remaining one was utilized for evaluation.
The proposed method achieved an average improvement of 1.36$\%$ in mIoU compared to the original U-Net when it served as the feature extractor, with a notable average increase of 2.71$\%$ in vascular regions. When we use UCTransNet as a feature extractor, the proposed method improved mIoU by 1.02$\%$ over the original UCTransNet, with a significant improvement of 2.17$\%$ in vascular regions.

Another experiment was conducted on 
MoNuSeg~\cite{MoNuSeg} dataset, which exhibits diversity in nuclei across multiple organs and patients and is captured under varying staining methods at different hospitals. Therefore, DG is required to extract features that are independent of domain differences.
Compared to the original U-Net~\cite{U-Net}, our method using U-Net as a feature extractor improved mIoU by 2.53$\%$, with an improvement of 2.73$\%$ in cell nuclei. Additionally, compared to UCTransNet~\cite{uc}, the proposed method using UCTransNet as a feature extractor also improved mIoU by 3.0$\%$, with an improvement of 3.65$\%$ in cell nuclei.

The structure of this paper is as follows. Section \ref{sec:2_related} describes related works. Section \ref{sec:3_proposed} explains the details of the proposed method. Section \ref{sec:4_experiments} presents and discusses the experimental results. Section \ref{sec:5_conclusion} describes conclusions and future work.

%% file: sec/2_related.tex
\section{Related Works}
\label{sec:2_related}

\subsection{Domain Generalization for Semantic Segmentation}

DG only allows access to the source domain and extracts features that are not specific to it (e.g., background or cell nuclei), mitigating domain shifts for unseen target domains.
Existing research on DG for semantic segmentation often focuses on methods that remove domain-specific style information. Techniques such as normalization~\cite{IBN,self-adaptation}, whitening~\cite{robustnet}, and diversification~\cite{style1,style2} have been used to achieve DG. However, WildNet argued that style information and content information are not orthogonal, and whitening style information can inadvertently remove necessary content information.
Therefore, to build a high-precision model, it is essential to effectively acquire information that is independent of the dataset. We propose a method that splits the feature maps from input images into two parts and trains them to decorrelate from each other. One feature map is constrained to acquire category information independent of domain differences, while the other retains source domain-specific information. This approach allows effective segmentation using feature maps that contain category information independent of domain differences.

\subsection{Image Generation Model}

Research on generative models has been extensive, with various approaches proposed, such as Variational Autoencoder (VAE)~\cite{VAE}, and Vector Quantized VAE (VQ-VAE)~\cite{VQVAE}.
VAE maps input data to a latent space as a probability distribution and samples latent variables from this distribution to generate images.
VQ-VAE possesses higher quality image generation capabilities and clustering abilities compared to VAE.
However, VQ-VAE has a non-differentiability issue due to the use of the argmax function for discretization. This problem was addressed by SQ-VAE.
SQ-VAE uses Gumbel-Softmax~\cite{gumbel} to approximate a categorical distribution in a differentiable manner, allowing uninterrupted backpropagation.
We propose a method focusing on the clustering capability of SQ-VAEs to bridge the domain gap between a source domain and an unseen target domain. Specifically, we constrain the probabilities to divide the N quantum vectors into K groups.
Aligning the groups for each category can minimize the gap with the unseen target domain. This approach helps mitigate the accuracy degradation caused by the target domain.

%% file: sec/3_method.tex
\section{Methodology}
\label{sec:3_proposed}

When conventional semantic segmentation learns from a specific dataset (source domain) and evaluates on an unseen dataset (target domain), accuracy decreases significantly due to domain differences such as different imaging devices and staining methods. To improve accuracy through DG, we propose two approaches.
First, we separate feature maps into domain-independent category information and domain-specific information of the source domain. Using domain-independent category information for segmentation, we believe that accuracy is independent of the dataset domain used for training.
Second, using only category information cannot completely bridge the domain gap between the source and target domains. Therefore, we propose a method to address the domain gap using the quantum vectors of SQ-VAE.

\subsection{Category Information Independent of Domain Differences} 
\label{sec:extcategory}

\begin{figure*}[t]
    \centering
    \includegraphics[width=0.9\linewidth]{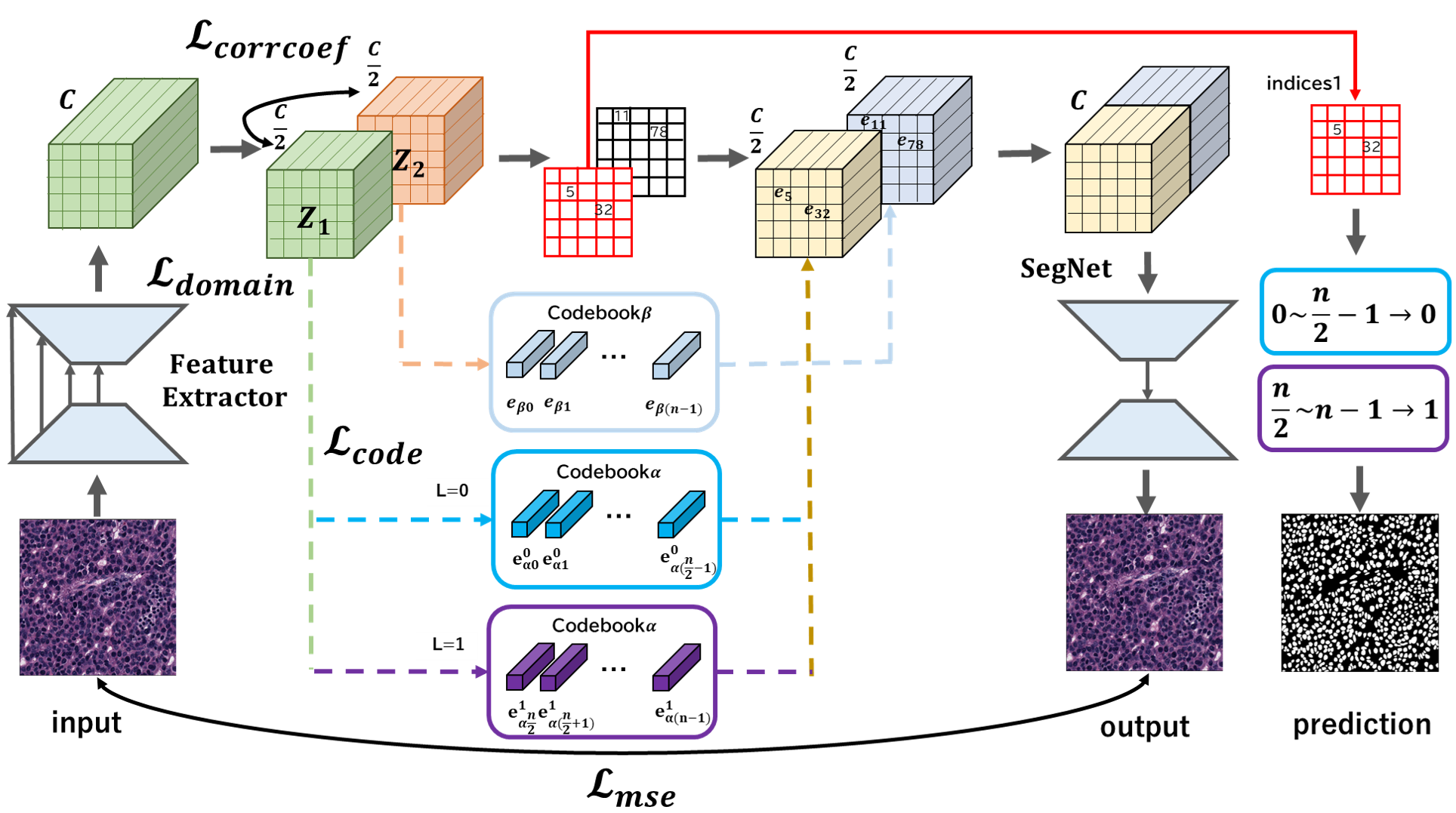}
    \caption{Overview of the proposed method. We extracted domain-independent category information to address unseen target domains. Domain-independent category information is represented using $Z_1$, which is used for segmentation. Additionally, quantum vectors are used for quantization to bridge the gap between source domain and unseen target domain.}
    \label{fig:proposed}
\end{figure*}

Figure \ref{fig:proposed} illustrates the overview of our method. To extract information independent of domain differences,
an input image $x\in \mathbb{R}^{3\times H \times W}$ is processed by a feature extractor, such as U-Net or UCTransNet,
which outputs the feature map $Z\in \mathbb{R}^{C\times H \times W}$. The feature extractor, such as U-Net or UCTransNet, achieves high accuracy and provides output images of the same dimensions as the input images. This is why we use U-Net or UCTransNet as the encoder. The feature maps obtained from the encoder are divided along the channel dimension into two parts: the domain-independent category information and the remaining information. The divided feature maps are denoted as $Z_1\in \mathbb{R}^{C_1\times H \times W}$ and $Z_2\in \mathbb{R}^{C_2\times H \times W}$ where $C_1 = C_2 = C / 2$.
To separate domain-independent category information from source domain-specific information, the model is trained to decorrelate the feature maps $Z_1$ and $Z_2$. In this paper, we adopt DeepCCA, which can learn nonlinear relationships, allowing it to handle more complex data and achieve high precision in removing correlations.

DeepCCA maximizes the correlation between two variables in a nonlinear manner. However, since our goal is to decorrelate them, squaring the output of DeepCCA brings the result closer to zero.
\begin{eqnarray}
  \mathcal{L}_{corrcoef} & = & corrcoef(Z_1, Z_2)^2
  \label{equation:1}
\end{eqnarray}
After decorrelation, feature maps are quantized using quantization vectors $e_\alpha \in \mathbb{R}^{N\times C_1}$ and $e_\beta \in \mathbb{R}^{N\times C_2}$.
$N$ is the dimension of the embedding vector space.
The details are presented in Section \ref{sec:SQ_seg}.
The reason for preparing $e_\alpha$ and $e_\beta$ is to assign them different roles. $e_\alpha$ is used to provide domain-independent categorical features, while $e_\beta$ supplies source domain-specific features.
This approach allows each to serve distinct functions. The feature maps after quantization are denoted as $Z_1^\prime$ and $Z_2^\prime$. $ Z_1^\prime$ and $Z_2^\prime$ are also trained to be decorrelated using DeepCCA.
\begin{eqnarray}
  \mathcal{L}_{corrcoef} & = & corrcoef(Z_1^\prime, Z_2^\prime)^2
  \label{equation:2}
\end{eqnarray}

By using constraints in Equations \ref{equation:1} and \ref{equation:2}, the quantization vectors $e_\alpha$ and $e_\beta$ become automatically uncorrelated. By training $Z_1$ and $Z_2$ to be uncorrelated, they can assume different roles.
For example, if we assume that $Z_1$ and $e_\alpha$ represent domain-independent category features, then $Z_2$ and $e_\beta$ represent source domain-specific features.
We hypothesize that using these domain-independent features for segmentation will improve accuracy. This method involves training multiple features to be uncorrelated, but there is no guarantee that $Z_1$ represents domain-independent category information.

To address this, we propose a method to group features within $Z_1$ that belong to the same category, embedding domain-independent category information. Specifically, we introduce a set of learnable representative vectors, $t^K\in \mathbb{R}^{K\times C_1}$. Here, $K$ denotes the number of categories. For example, for category $t^0$, we train the model to cluster features $Z_{1}^0$ in $Z_1$ with a target label of 0. This process is repeated for all $K$ categories. Additionally, to ensure that these representative vectors do not capture source domain-specific information, we train the model to separate the representative vectors from each other. By doing so, we can densely embed domain-independent category information.
\begin{eqnarray}\label{domain}
\mathcal{L}_{domain} = \sum_{i=0}^{K-1}\Big\{||{t}^{i} - {Z}_{1}^i||_{2}\ - \sum_{j=0}^{K-1}\frac{||t^i - t^j||_2}{2}\Big\}
\label{equation:3}
\end{eqnarray}
where $Z_{1}^0$ and $Z_{1}^1$ are defined as the features of $Z_1$ when the target label is 0 and 1.

\subsection{Segmentation using SQ-VAE} 
\label{sec:SQ_seg}

In Section \ref{sec:extcategory}, we proposed a method that divides features into those related to domain-independent categories and those specific to the source domain. Although only domain-independent features provide DG capabilities, this does not completely prevent the domain gap. Thus, we propose a method to bridge the domain gap between the source and unseen target domains using quantum vectors from the SQ-VAE. The reason for using SQ-VAE is that it can generate various clusters by utilizing quantum vectors. Additionally, SQ-VAE is more accurate as an image generation model compared to VAE and VQ-VAE.
We constrain the $N$ quantum vectors from the SQ-VAE into $K$ groups to address the domain gap.
For instance, in the category of cell nuclei, we train the model to bring the group of quantum vectors representing cell nuclei closer to the group of features obtained from input images of cell nuclei. By aligning these groups, we can mitigate the domain gap between the source and unseen target domains, allowing for better generalization.
We feed an image $x\in \mathbb{R}^{3\times H \times W}$ into the feature extractor, such as U-Net and UCTransNet, which outputs a feature map $Z\in \mathbb{R}^{C\times H \times W}$. As explained in Section \ref{sec:extcategory}, to separate domain-independent category features from domain-specific features, we divide the feature map into two parts: $Z_1$ and $Z_2$.

To address the domain gap, we define the quantum vectors as $e\in \mathbb{R}^{N\times C/2}$ where $C/2$ is the channel dimension of the embedded vector $e$. We separate domain-independent category information from domain-specific in ${e_\alpha}\in \mathbb{R}^{N\times C_1}$ and ${e_\beta}\in \mathbb{R}^{N\times C_2}$. To quantize the feature map obtained from the feature extractor, we calculate the Mahalanobis distance between the feature map $Z_1$ and the quantum vector $e_\alpha$.
\begin{eqnarray}\label{logit1}
  {logit}_1 & = & - \Big\{ \frac{({e}_{\alpha j}-{Z}_{1})^\top \Sigma_{\gamma}^{-1} ({e}_{\alpha j}-{Z}_{1}) }{2} \Big\}_{j=0}^{N-1}
\end{eqnarray}
where $\Sigma_{\gamma}$ is a learnable parameter, and $j$ refers to one of the quantum vectors within the set of $N$ vectors.
It is denoted as $\sum_{\gamma} = \sigma_{\gamma}^{2}{I}$.
Then, ${logit}_1$ is expressed as
\begin{eqnarray}\label{logit1-l2}
  {logit}_1 & = & \frac{||{e}_{\alpha j}-{Z}_{1}||_{2}^{2}}{2\sigma_{\gamma}^{2}}
\end{eqnarray}
where $logit_1\in \mathbb{R}^{HW\times N}$ is a matrix. In this case, $\sum_{\gamma}$ = $\sigma_{\gamma}^{2}{I}$ is learned to approach zero from the initial value. As training progresses, the probabilities of the distances between feature maps obtained by the encoder and the quantum vectors become closer to a one-hot encoding. This is similar to SQ-VAE.
To probabilistically quantify the Mahalanobis distance between the obtained feature map and the quantum vectors, we use Gumbel-Softmax. We use Gumbel-Softmax because it approximates the selection of discrete quantum vectors as a continuous probability distribution and is differentiable.
\begin{eqnarray}
  \label{p1}
  {P}_1 & = &{\text{Gumbelsoftmax}\left( - \frac{{logit}_1}{\tau} \right)}
\end{eqnarray}
where $\tau$ is a learnable temperature parameter. Similarly, we use $Z_2$ and $e_\beta$ to output $P_2$. As shown in Equations \ref{logit1-l2} and \ref{p1}, we calculate the Mahalanobis distance and convert it to probabilities using Gumbel-Softmax. These probabilities are then used to quantize the features obtained from the encoder.

During evaluation, we replace the feature map obtained from Equation \ref{logit1-l2} with the quantum vector that has the closest Mahalanobis distance.
\begin{eqnarray}
  indices_{1} & = & {\text{argmax}(logit_{1})}
\end{eqnarray}
where argmax is used along the N-dimensional direction of $logit_1\in \mathbb{R}^{HW\times N}$.

The feature maps fed into the decoder are quantized as  ${Z}^{\prime}_{1}\in \mathbb{R}^{{C}_1 \times H \times W}$ and ${Z}^{\prime}_{2}\in \mathbb{R}^{{C}_2 \times H \times W}$. Since the channel dimension is split into two, the dimensions from $Z_1^{\prime}$ and $Z_2^{\prime}$ are combined.
\begin{eqnarray}\label{concat}
  {Z}^{\prime} & = & {\text{Concat}({Z}^{\prime}_{1}, {Z}^{\prime}_{2})}
\end{eqnarray}
According to Equation \ref{concat}, it becomes ${Z}^{\prime}\in \mathbb{R}^{C \times H \times W}$. The decoder shown in Figure \ref{fig:proposed} uses an encoder-decoder CNN ~\cite{SegNet} without skips. This is because U-Net or UCTransNet features transmission mechanisms that allow the input image to flow through easily, simplifying reconstruction and hindering the learning of intermediate layers. The input ${Z}^{\prime}$ is passed through the encoder-decoder CNN, producing the final output $x^{\prime}$. The reconstruction error between the input image $x$ and the final output image $x^{\prime}$ is then calculated.
\begin{eqnarray}
  \mathcal{L}_{mse} & = & \frac{\sum_{i=1}^{n}\log {(x_i - x^{\prime}_i)^2}}{2}
\end{eqnarray}
where $n$ is the total number of pixels in the input image. The reason for adding $\log$ to the loss is that the gradient becomes larger as $x$ decreases. In other words, we believe that the model should focus on finer details during reconstruction. This is consistent with the implementation of SQ-VAE.

The objective is to use features related to categories that are independent of domain differences and employ quantum vectors to prevent domain gaps for unseen target domains. Additionally, segmentation is performed by the indices of the quantum vectors. This is executed using $Z_1$ and $e_\alpha \in \mathbb{R}^{N\times C_1}$.
It is divided into K parts to separate roles based on the number of segmentation categories $K$.

In this paper, we consider the case where K=2. First, it is divided into two parts: $e_{\alpha}^0 \in \mathbb{R}^{N_1\times C_1}$, corresponding to category label 0, and $e_{\alpha}^1 \in \mathbb{R}^{N_2\times C_1}$, corresponding to category label 1. Let $N_1$ and $N_2$ be such that $N_1 = N_2 = N/2$. From Equation \ref{p1}, we have $P_1\in \mathbb{R}^{H\times W\times N}$. When the category label is 0, we want the probability of selecting $e_{\alpha}^0$ to be 1. Similarly, when the category label is 1, we want the probability of selecting $e_{\alpha}^1$ to be 1. To achieve this, we train the model such that the sum of probabilities in the $N_1$ dimensional direction of $P_1$ is 1 when the category label is 0, and the sum of probabilities in the $N_2$ dimensional direction of $P_1$ is 1 when the category label is 1.
In other words, when the category label is 0, the model is trained to minimize the distance between the feature map $Z_1^0\in \mathbb{R}^{C_1\times H_0 \times W_0}$ related to the category label and $e_{\alpha}^0 \in \mathbb{R}^{N_1\times C_1}$ where $H_0$ and $W_0$ correspond to category label 0.
When the category label is 1, the model is trained to minimize the distance between the feature map $Z_1^1\in \mathbb{R}^{C_1\times H_1 \times W_1}$ related to the category label and $e_{\beta}^1 \in \mathbb{R}^{N_2\times C_1}$ where $H_1$ and $W_1$ correspond to category label 1. This learning method is as
\footnotesize
\begin{eqnarray}\label{code}
  \mathcal{L}_{\text{code}} = \log \Big\{ w_0 \times(1 - \sum_{N=0}^{\frac{N}{2}-1}P_{1})^2_{\text{L}=0} \nonumber \\ + w_1 \times(1 - \sum_{N=\frac{N}{2}}^{N-1}P_{1})^2_{\text{L}=1}\Big\}
\end{eqnarray}
\normalsize
where $L$ is the label, and $w_0$ and $w_1$ are weights added to prioritize uncertain predictions. By using these weights, the model can focus on learning parts where predictions are uncertain, such as the boundaries in segmentation.

\begin{figure}[t]
    \centering
    \includegraphics[width=0.7\linewidth]{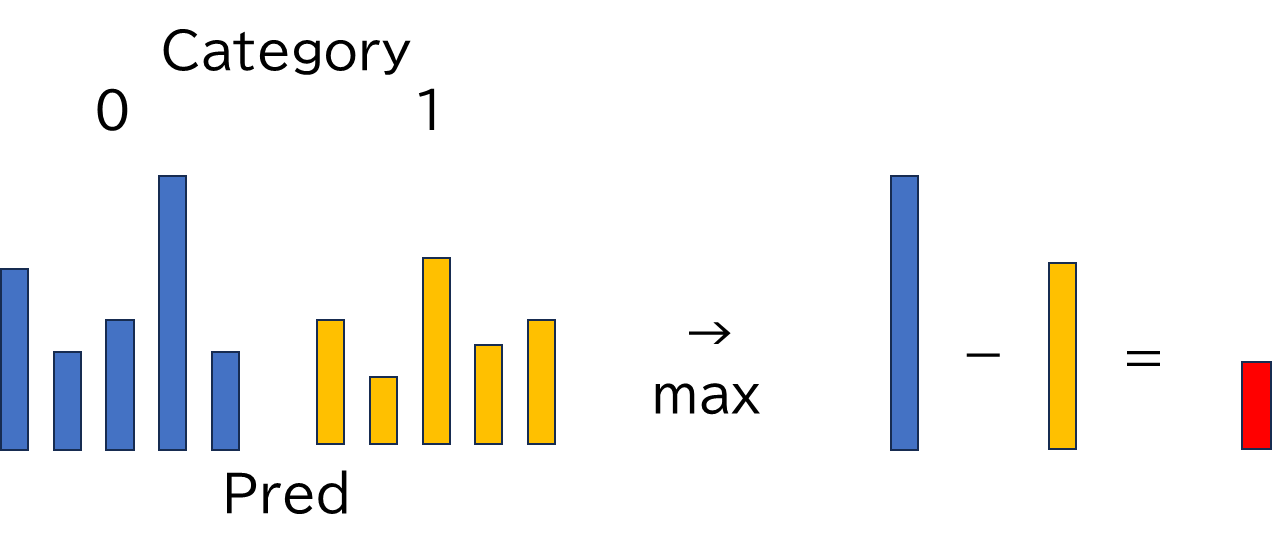}
    \caption{Weights that are learned to focus on parts where predictions become uncertain. Treat red as weight.}
    \label{fig:weight}
\end{figure}
Figure \ref{fig:weight} shows a conceptual diagram of the weights. The maximum values of the prediction probabilities from index 0 to (N-1)/2 and from N/2 to N-1 are obtained. These are defined as $max(P^0_1)$ and $max(P^1_1)$, respectively. Next, the absolute value of the difference between these maximum values is taken as
\begin{eqnarray}\label{dif}
  dif & = & |max(P_1^0) - max(P_1^1)|
\end{eqnarray}
The absolute differences are collected for all pixels and divided by the maximum value among all pixels to normalize them from 0 to 1. The weight of the i-th pixel is as follows:
\begin{eqnarray}\label{weight}
  w_i & = & \frac{dif_i}{max(dif)}
\end{eqnarray}
The weight $w$ is less likely to be learned if it is closer to 1, as it indicates higher certainty. Conversely, if it is closer to 0, it signifies greater uncertainty in the prediction, and it is learned more intensively.
The reason weights closer to 0 are learned more intensively than those closer to 1 is that Equation \ref{code} includes a $\log$, making smaller loss values more significant. Taking the derivative of $\log(x)$ results in $1/x$, meaning that as $x$ becomes smaller, the gradient becomes larger, thereby making smaller loss more significant.

Finally, the index of the quantization vector closest to the feature vector is obtained, which is related to the category and independent of the domain differences. If this index is between $0$ and $(N/2)-1$, category label 0 is assigned. If the index is between $N/2$ and $N-1$, category label 1 is assigned. This category label is used as the final segmentation prediction. The learning method using this quantized vector is shown in Figure \ref{fig:domaincodebook}, with Step 2 particularly pertaining to that part.
When the category label is 0, the model is trained so that the sum in the $N_1$ dimensional direction of $P_1$ equals 1. When the category label is 1, the model is trained so that the sum in the $N_2$ dimensional direction of $P_1$ equals 1. This approach divides the features into two groups using the labeled data, similar to dividing the quantization vectors into two groups. The model is then trained to bring the groups closer together. As a result, even in the presence of a domain gap in the unseen target domain, it can be mitigated, as quantization assigns similar quantum vectors to clusters (groups of the same category).

%% file: sec/4_experiments.tex
\section{Experiments}
\label{sec:4_experiments}

\subsection{Implementation Details}

Our method is evaluated on blood vessel segmentation from three types of fundus images: Drive, Stare, and Chase. The Drive, Stare, and Chase datasets each contain a total of 20, 20, and 28 images, respectively, along with annotations for segmenting the images into classes: background and blood vessels.
The images in the Drive dataset were captured using a Canon CR5 non-mydriatic 3CCD camera. The images in the Stare dataset were obtained using a Top Con TRV-50 retinal camera. The images in the Chase dataset were captured with a Nidek NM-200-D fundus camera. As these images were taken with different cameras,
they can be considered to belong to different domains.

Of the three datasets, two are used for training, while the remaining dataset is used for evaluation. By rotating this arrangement, the DG performance is assessed.
Each dataset is divided into five parts to perform 5-fold cross-validation, with four parts used for training (e.g., 4/5 of Drive and 4/5 of Chase) and one part used for validation (e.g., 1/5 of Drive and 1/5 of Chase).
Subsequently, the two datasets used for training and validation are combined to ensure there is no data imbalance.
The validation data from the remaining dataset, which was not used for training, is used for evaluation (e.g., 1/5 of Stare). However, since the Chase dataset contains more images, to avoid bias in training or evaluation, the number of images in the Chase dataset is randomly reduced to 20 for the experiments.

Additionally, we conduct experiments on the MoNuSeg dataset, which contains tissue images of tumors from various organs diagnosed in several patients across multiple hospitals.
Due to the diverse appearance of nuclei across different organs and patients, as well as the variety of staining methods used by various hospitals, it is important to extract domain-agnostic information from this dataset.
The MoNuSeg dataset consists of 30 images for training and 14 images for evaluation. Among the training data, 24 images are allocated for training, and 6 images are reserved for validation. The test data is used as is for evaluation and includes lung and brain cells that are not present in the training data, rendering them unseen data.


For all experiments, the seed value is changed four times to calculate average accuracy. We resize all images to $256\times256$ pixels as preprocessing. The learning rate is set to $1\times 10^{-3}$, the batch size is $2$, the optimizer is Adam, and the number of epochs is $200$. We used an Nvidia RTX A6000 GPU. The number of quantum vectors is set to $512$. The evaluation metric is intersection over union (IoU), and we evaluate using the IoU for each class and the mean IoU (mIoU) across all classes. We compared the proposed method with U-Net and UCTransNet. The rationale is that the proposed method uses U-Net or UCTransNet as an encoder and makes predictions using quantum vectors based on its output. In other words, the same feature extractor is used up to the point of segmentation prediction.

\subsection{Domain Generalization on Chase, Stare, and Drive Datasets}

The results of DG on the Chase, Stare, and Drive datasets are shown in Table \ref{table:vessel}. The method with the highest accuracy is shown in orange, while the second-highest accuracy is in blue.
When the Drive and Stare datasets were used for training and the Chase dataset for evaluation, the proposed method (U-Net+ours) using U-Net as a feature extractor exhibited a 1.80$\%$ improvement in mIoU compared to the original U-Net, with a specific improvement of 3.69$\%$ in the blood vessel area. Additionally, the proposed method (UCTransNet+ours) using UCTransNet as a feature extractor demonstrated a 2.41$\%$ improvement in mIoU compared to the original UCTransNet, with a specific improvement of 5.01$\%$ in the blood vessel area.
When the Drive and Chase datasets were used for training and the Stare dataset for evaluation, the proposed method (U-Net+ours) using U-Net as a feature extractor showed a 1.20$\%$ improvement in mIoU compared to the original U-Net, with a specific improvement of 2.47$\%$ in the blood vessel area. Additionally, the proposed method (UCTransNet+ours) using UCTransNet as a feature extractor noted a 0.32$\%$ improvement in mIoU compared to the original UCTransNet, with a specific improvement of 0.76$\%$ in the blood vessel area.
When the Stare and Chase datasets were used for training and the Drive dataset for evaluation, the proposed method (U-Net+ours) using U-Net as a feature extractor showed a 1.09$\%$ improvement in mIoU compared to the original U-Net, with a specific improvement of 1.98$\%$ in the blood vessel area. Additionally, the proposed method (UCTransNet+ours), using UCTransNet as a feature extractor showed a 0.35$\%$ improvement in mIoU compared to the original UCTransNet, with a specific improvement of 0.75$\%$ in the blood vessel area.
These improvements indicate that DG is effectively achieved.

\begin{figure*}[t]
    \centering
    \includegraphics[width=0.79\linewidth]{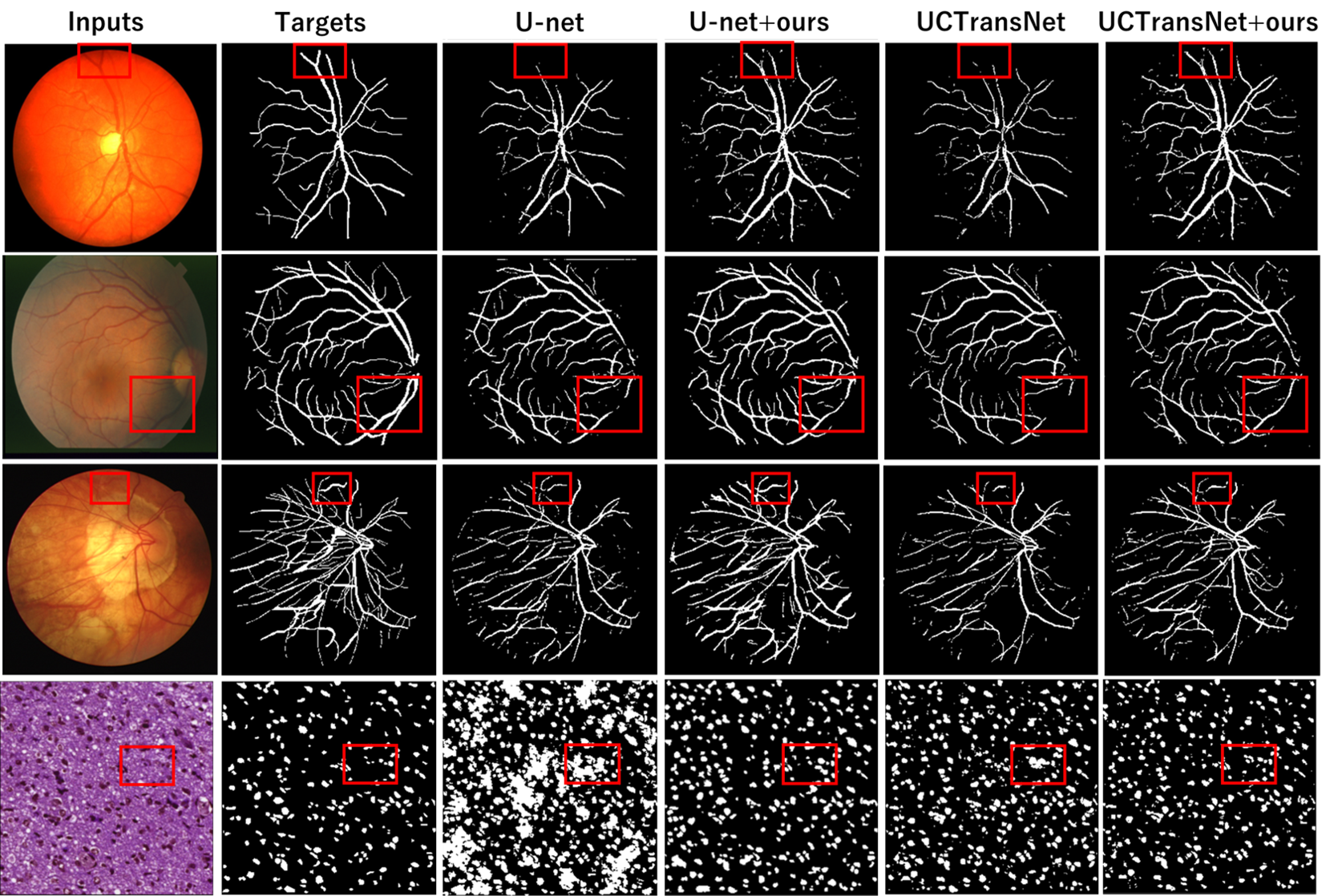}
    \caption{Segmentation results on Chase, Stare, MoNuSeg, and Drive datasets. From left to right, the images show input images, ground truth, results by original U-Net, our method (U-Net+ours), original UCTransNet, and our method (UCTransNet+ours).}
    \label{fig:visual}
\end{figure*}

Additionally, segmentation results are shown in Figure \ref{fig:visual}.
The top three rows display the results on the Chase, Stare, and Drive datasets.
The areas highlighted in red boxes show significant improvements. For the Chase dataset, vascular regions in the red box of the input image appear slightly darker, which the original U-Net and UCTransNet predict as background. In contrast, our methods (U-Net+ours and UCTransNet+ours) extract category information independently of the domain, preventing domain gaps, can densely extract blood vessel category information, predicting them correctly. For the Stare dataset, focusing on the red box areas, the original U-Net and UCTransNet make predictions indicating disconnected blood vessels. However, the proposed methods (U-Net+ours and UCTransNet+ours) predict connected blood vessels, effectively extracting category information independently of the domain. For the Drive dataset, in the red box areas, the original U-Net and UCTransNet predict the blood vessels as thin or disconnected. In contrast, the proposed methods predict thicker blood vessels and connect previously disconnected vessels, successfully extracting domain-independent information.

\begin{table}[t]
\caption{IoU and standard deviation Chase, Stare, and Drive datasets. \textcolor{orange}{orange} indicates the highest accuracy, and \textcolor{blue}{blue} indicates the second-highest accuracy.}
\label{table:vessel}
\centering
\scalebox{0.65}{
\begin{tabular}{c|c|ccc}
\hline
datasets & methods            & background          & blood vessels        & mIoU                \\
\hline
\hline
\multirow{4}{*}{Chase}
 & U-Net       & \textcolor{orange}{95.33}($\pm$0.35)  & 43.56($\pm$4.14) & 69.44($\pm$2.15) \\
 & U-Net + ours        & 95.24($\pm$0.48)  & \textcolor{blue}{47.25}($\pm$1.82) & \textcolor{blue}{71.24}($\pm$1.13)  \\ 
 & UCTransNet   & \textcolor{blue}{95.27}($\pm$0.37)  & 45.92($\pm$3.84) & 70.59($\pm$1.97) \\
 & UCTransNet + ours        & 95.06($\pm$0.39)  & \textcolor{orange}{50.93}($\pm$1.65) & \textcolor{orange}{73.0}($\pm$0.92)  \\
\hline
\hline
\multirow{4}{*}{Stare}
 & U-Net       & \textcolor{orange}{95.81}($\pm$0.86)  & 56.70($\pm$6.79) & 76.26($\pm$3.73) \\
 & U-Net + ours        &  95.75($\pm$0.71) & \textcolor{orange}{59.17}($\pm$4.16) & \textcolor{orange}{77.46}($\pm$2.36) \\ 
 & UCTransNet   &  \textcolor{blue}{95.80}($\pm$0.85)  & 56.71($\pm$5.66) & 76.25($\pm$3.20) \\
 & UCTransNet + ours   &  95.67($\pm$0.71)  & \textcolor{blue}{57.47}($\pm$4.17) & \textcolor{blue}{76.57}($\pm$2.36) \\
\hline
\hline
\multirow{4}{*}{Drive}
 & U-Net   & 95.56($\pm$0.53)  & 57.86($\pm$1.85) & 76.71($\pm$1.17) \\
 & U-Net + ours    & \textcolor{orange}{95.75}($\pm$0.40) & \textcolor{orange}{59.84}($\pm$3.06) & \textcolor{orange}{77.80}($\pm$1.57) \\
 & UCTransNet   & \textcolor{blue}{95.62}($\pm$0.49)  & 58.35($\pm$2.10) & 76.99($\pm$1.28) \\
 & UCTransNet + ours   & 95.58($\pm$0.46)  & \textcolor{blue}{59.10}($\pm$1.58) & \textcolor{blue}{77.34}($\pm$1.01) \\
\hline
\end{tabular}
}

\end{table}

\subsection{Domain Generalization on MoNuSeg}

\begin{table}[t]
\centering
\caption{IoU and standard deviation on MoNuSeg dataset. \textcolor{orange}{orange} method achieved the highest accuracy, while \textcolor{blue}{blue} method attained the second-highest accuracy.}
\label{table:monuseg}
\scalebox{0.62}{
\begin{tabular}{c|c|ccc}
\hline
dataset & methods & background & cell nucleus & mIoU \\ \hline
\hline
\multirow{4}{*}{MoNuSeg}
& U-Net & 87.36 ($\pm$ 0.96) & 61.55 ($\pm$ 1.17) & 74.46 ($\pm$ 1.01) \\
& U-Net + ours & \textcolor{blue}{89.71} ($\pm$ 1.24) & \textcolor{blue}{64.28} ($\pm$ 1.10) & \textcolor{blue}{76.99} ($\pm$ 1.17) \\
& UCTransNet & 87.79 ($\pm$ 0.07) & 60.93 ($\pm$ 0.51) & 74.36 ($\pm$ 0.23) \\
& UCTransNet + ours & \textcolor{orange}{90.15} ($\pm$ 0.95) & \textcolor{orange}{64.58} ($\pm$ 0.67) & \textcolor{orange}{77.36} ($\pm$ 0.80) \\
\hline
\end{tabular}
}
\end{table}

The results of DG on the MoNuSeg dataset are shown in Table \ref{table:monuseg}. The method with the highest accuracy is highlighted in orange, while the method with the second-highest accuracy is shown in blue. Comparison results between conventional methods and the proposed methods (methods + ours) are included. As a result, the proposed method (U-Net+ours) achieved a 2.53$\%$ improvement in mIoU compared to the original U-Net, with a specific improvement of 2.73$\%$ in the cell nucleus area. Additionally, the proposed method (UCTransNet+ours) achieved a 3.0$\%$ improvement in mIoU compared to the original UCTransNet, with a specific improvement of 3.65$\%$ in the cell nucleus area. These results demonstrate strong generalization performance to unseen target domains. The improved accuracy in cell nuclei, in particular, suggests that the method can effectively handle the diversity in the appearance of cell nuclei and staining methods across different domains, successfully extracting cell nucleus-specific features.

Segmentation results from conventional methods and our methods are displayed in the bottom row of Figure \ref{fig:visual}.
The areas highlighted in red boxes indicate where significant improvements were observed.
The original U-Net struggled with DG, often predicting background as cell nuclei. In contrast, the proposed method (U-Net+ours) successfully extracted category information independent of the domain, closing domain gaps and enabling accurate predictions. Similarly, the original UCTransNet over-predicted cell nuclei in the areas highlighted in red. However, the proposed method (UCTransNet+ours) effectively extracted information on various cell nuclei, leading to accurate predictions.

%% file: sec/5_conclusion.tex
\section{Conclusion}
\label{sec:5_conclusion}
We proposed a method that generalizes well to datasets with differences in imaging equipment and staining methods (target domain) compared to the dataset (source domain) on which the model was trained. The proposed method showed significant improvements on cell image datasets with various staining methods and fundus images captured by different imaging devices. These results demonstrate the generalization performance of our method to unseen target domains. In the future, we would like to evaluate our method on a multi-class segmentation problem.